\documentclass[letterpaper]{article} % DO NOT CHANGE THIS
\usepackage{aaai2026}  % DO NOT CHANGE THIS
\usepackage{times}  % DO NOT CHANGE THIS
\usepackage{helvet}  % DO NOT CHANGE THIS
\usepackage{courier}  % DO NOT CHANGE THIS
\usepackage[hyphens]{url}  % DO NOT CHANGE THIS
\usepackage{graphicx} % DO NOT CHANGE THIS
\usepackage{natbib}  % DO NOT CHANGE THIS AND DO NOT ADD ANY OPTIONS TO IT
\usepackage{caption} % DO NOT CHANGE THIS AND DO NOT ADD ANY OPTIONS TO IT
\usepackage{algorithm}
\usepackage{algorithmic}
\usepackage{amsmath}
\usepackage{amsfonts}

\usepackage{newfloat}
\usepackage{listings}
\usepackage{tcolorbox}
\usepackage{multirow}
\usepackage{subcaption}
\newcommand{\code}[1]{{\ttfamily#1}}
\usepackage{newfloat}
\usepackage{listings}
\DeclareCaptionStyle{ruled}{labelfont=normalfont,labelsep=colon,strut=off} % DO NOT CHANGE THIS
\floatstyle{ruled}
\newfloat{listing}{tb}{lst}{}
\floatname{listing}{Listing}
\title{Automatic Paper Reviewing with Heterogeneous Graph Reasoning over LLM-Simulated Reviewer-Author Debates}
\author{
    Shuaimin Li \textsuperscript{\rm 1,} \equalcontrib, Liyang Fan\textsuperscript{\rm 2, 3,} \equalcontrib, Yufang Lin\textsuperscript{\rm 4,} \equalcontrib, Zeyang Li\textsuperscript{\rm 5}, Xian Wei\textsuperscript{\rm 4}, Shiwen Ni \textsuperscript{\rm 3,} \thanks{Corresponding authors.}\\ Hamid Alinejad-Rokny \textsuperscript{\rm 6}, Min Yang \textsuperscript{\rm 1, 3,} $^\dagger$
}
\affiliations{
    \textsuperscript{\rm 1}Shenzhen Key Laboratory for High Performance Data Mining, Shenzhen Institutes of Advanced Technology, Chinese Academy of Sciences, Shenzhen, China \\
    \textsuperscript{\rm 2} College of Computer Science and Software Engineering, Shenzhen University, Shenzhen, China \\
    \textsuperscript{\rm 3} Artificial Intelligence Research Institute, Shenzhen University of Advanced Technology, Shenzhen, China\\
    \textsuperscript{\rm 4} East China Normal University, Shanghai, China\\
    \textsuperscript{\rm 5} University of Science and Technology of China, Suzhou, China\\
    \textsuperscript{\rm 6} University of New South Wales, Sydney, New South Wales, Australia\\sm.li2@siat.ac.cn, 2410833006@mails.szu.edu.cn, 71265902002@stu.ecnu.edu.cn, zeyangli@163.com,\\xian.wei@tum.de, sw.ni@siat.ac.cn,h.alinejad@unsw.edu.au, min.yang@siat.ac.cn

}

\usepackage{bibentry}
\begin{document}

\maketitle

\begin{abstract}
Existing paper review methods often rely on superficial manuscript features or directly on large language models (LLMs), which are prone to hallucinations, biased scoring, and limited reasoning capabilities. Moreover, these methods often fail to capture the complex argumentative reasoning and negotiation dynamics inherent in reviewer-author interactions. To address these limitations, we propose \textbf{ReViewGraph} (Reviewer-Author Debates Graph Reasoner), a novel framework that performs heterogeneous graph reasoning over LLM-simulated multi-round reviewer-author debates. In our approach, reviewer-author exchanges are simulated through LLM-based multi-agent collaboration. Diverse opinion relations (e.g., acceptance, rejection, clarification, and compromise) are then explicitly extracted and encoded as typed edges within a heterogeneous interaction graph. By applying graph neural networks to reason over these structured debate graphs, ReViewGraph captures fine-grained argumentative dynamics and enables more informed review decisions. Extensive experiments on three datasets demonstrate that ReViewGraph outperforms strong baselines with an average relative improvement of 15.73\%, underscoring the value of modeling detailed reviewer–author debate structures.

\end{abstract}

\begin{links}
    \link{Code}{https://github.com/relic-yuexi/ReViewGraph}

\end{links}

\section{Introduction}
Peer review is essential to scientific progress, ensuring the quality, validity, and originality of research~\cite{alberts2008reviewing}. High-quality reviews help guide the research community toward impactful contributions. However, the recent surge in paper submissions across disciplines~\cite{drozdz2024peer} has placed increasing strain on the peer review process. This exponential growth imposes a significant burden on human reviewers, making it increasingly difficult to ensure timely and consistent evaluations~\cite{StelmakhSSD21}. Moreover, peer review is inherently subjective. In some cases, reviewers may be careless, biased, or even malicious, thereby undermining the fairness and reliability of the decision-making process~\cite{StelmakhSSD21,fairness-peer-review}. These challenges have motivated the development of automatic reviewing systems as a promising solution to alleviate reviewer workload and promote more objective evaluations at scale.

With the rapid advancement of large language models (LLMs), the field of automatic peer review has made notable progress in recent years. In the computer science community, several top-tier conferences have begun experimenting with LLM-based reviewing systems as supplementary tools to assist human reviewers. For instance, ICLR 2025 introduced an LLM-powered automatic review system to complement human assessments\footnote{\url{https://blog.iclr.cc/2024/10/09/iclr2025-assisting-reviewers/}}, while AAAI 2026 has officially announced plans to integrate AI-assisted reviewing, using LLM-generated opinions as an additional perspective for expert evaluation\footnote{\url{https://aaai.org/conference/aaai/aaai-26/main-technical-track-call/}}.

Current LLM-based automatic reviewing methods generally fall into two main categories: (1) Prompt-based approaches, which leverage the in-context learning capabilities of LLMs to generate review content without parameter updates. These include direct prompting methods (e.g., AI-Scientist~\cite{ai-scientist}, which uses GPT-4 with tailored review templates) and multi-agent collaboration frameworks, where multiple LLM instances simulate interactions between reviewers and authors (e.g., AgentReview~\cite{agent-review} and ReviewMT~\cite{reviewMT}). (2) Fine-tuned approaches, which train open-source LLMs (e.g., LLaMA~\cite{llama} or Qwen~\cite{qwen} series) on peer review data to align the model's outputs with expert reviewing criteria (e.g., CycleReviewer~\cite{weng2025cycleresearcher} and DeepReview~\cite{zhu2025deepreviewimprovingllmbasedpaper}).

Despite recent progress in automatic paper reviewing, existing methods still face several notable challenges. Prompt-based approaches, which rely solely on LLMs and instructions, often generate superficial and shallow review content~\cite{ZhouC024}. Studies have shown that LLMs tend to produce generally positive but low-discriminative evaluations, failing to capture the nuanced reasoning required in peer review~\cite{grapheval}. Moreover, these methods are highly sensitive to prompt design~\cite{ErricaSSB25}; small changes in prompt wording can lead to significantly different outputs, resulting in limited stability and robustness. On the other hand, fine-tuned approaches that adapt open-source LLMs using peer review data suffer from data scarcity and bias. Publicly available high-quality review datasets are limited in size and scope, which restricts the generalizability of trained models. Additionally, these methods typically produce only a single-perspective review, lacking the ability to model multi-reviewer interactions and argumentative dynamics inherent in real-world peer review. Furthermore, both categories remain susceptible to hallucinations, wherein LLMs generate factually incorrect or misleading content, compromising fairness and reliability~\cite{HuangYMZFWCPFQL25}.

To address these challenges, we propose \textbf{ReViewGraph}, a novel framework for automatic paper reviewing that performs heterogeneous graph reasoning over LLM-simulated reviewer-author debates. Our approach explicitly captures multi-perspective opinions and their discourse-level interactions, thereby enhancing the depth, interpretability, and controllability of automated reviewing.

\textbf{ReViewGraph} begins by simulating multi-round reviewer-author debates through a multi-agent collaboration framework, resulting in a rich set of opinion exchanges. We then construct a heterogeneous debate graph to represent these interactions, comprising four node types: \textit{Title}, \textit{Evaluation Dimension}, \textit{Reviewer Opinion}, and \textit{Author Opinion}, and four meta-relation types that encode review structure and argumentative dynamics. These include (1) paper-to-dimension associations, (2) reviewer opinions tied to specific evaluation criteria (e.g., \textit{Methodological Novelty}, \textit{Motivation Clarity}, \textit{Experimental Completeness}, \textit{Writing Fluency}), (3) inter-reviewer relations such as \textit{agree}, \textit{disagree}, and \textit{complement}, and (4) reviewer-author interactions like \textit{clarify}, \textit{reject}, and \textit{accept}. To instantiate this graph, we use in-context prompting to extract opinion triplets and classify opinions into evaluation dimensions. This structured representation enables ReViewGraph to perform fine-grained relational reasoning and make informed final decisions. Building on this structured graph, we further apply a heterogeneous graph Transformer to perform relational reasoning and predict the final review decision.

To evaluate the effectiveness of ReViewGraph for automatic paper reviewing, we collect three benchmark datasets from OpenReview and compare our approach against seven strong baseline methods. Experimental results demonstrate that ReViewGraph consistently outperforms all baselines across these datasets. Notably, it achieves an average relative improvement of 15.73\% over the second-best baselines. Overall, the main contributions of this work are:
(1) We propose \textbf{ReViewGraph}, a novel framework for automatic paper reviewing that models reviewer–author interactions as heterogeneous graphs constructed from LLM-simulated multi-round debates.
(2) We design a structured heterogeneous debate graph with semantically-typed nodes and edges to capture fine-grained argumentative relations across diverse review perspectives, and leverage a graph neural network to perform relational reasoning over this structure.
(3) Extensive experiments on three datasets show that ReViewGraph consistently outperforms 7 strong baselines, achieving an average relative improvement of 15.73\% over the second-best models.

\section{Related Works}
\subsubsection{Traditional Automatic Reviewing.} Early efforts in automatic paper reviewing primarily relied on manually curated review data and traditional neural classifiers for acceptance prediction. \citet{KangADZKHS18} introduced the PeerRead dataset and trained classifiers using manually labeled reviews. \citet{GhosalVEB19} incorporated sentiment features to improve prediction accuracy. To move beyond simple acceptance prediction, researchers began exploring automatic review generation. \citet{BartoliLMT16} proposed one of the earliest neural frameworks, trained on 48 papers from their lab, to generate review comments. \citet{Nagata19} generated sentence-level feedback on grammar errors to support academic writing. ReviewRobot~\cite{WangZHKJR20} constructed a knowledge graph from the input paper, predicted review scores, and selected templated comments based on both scores and supporting evidence. However, traditional neural models struggled with long and technical documents, lacking the capacity for deep semantic understanding, which ultimately hindered progress in this field.

\subsubsection{LLM-based Automatic Reviewing.}
Recent advancements in LLMs have inspired a growing body of work on LLM-based automatic reviewing. Multi-agent collaboration frameworks aim to simulate the multi-role dynamics of real-world peer review. For example, ReviewMT~\cite{reviewMT} and AIScientist~\cite{ai-scientist} reframe the review process as a multi-round, long-context dialogue among multiple roles, i.e., reviewers, authors, and meta-reviewers. AgentReview~\cite{agent-review} further explores this direction by prompting LLMs to assume diverse reviewer personalities and decision strategies. 
Fine-tuned reviewer models enhance alignment with human review standards through supervised training. CycleReviewer~\cite{weng2025cycleresearcher} fine-tunes an open-source LLM on domain-specific review data, simulating multiple reviewers who assess the paper across different dimensions. DeepReview~\cite{zhu2025deepreviewimprovingllmbasedpaper} extends this line of work by modeling multi-step reviewer reasoning and training on generated rationales to produce more coherent, logically grounded reviews. Graph-based approaches such as GraphEval~\cite{grapheval} leverage LLM prompting to segment abstracts into discrete opinion sentences, which are then connected via similarity-based edges and processed through graph reasoning to predict acceptance decisions.

While these methods represent meaningful progress, they either treat interactions implicitly or lack fine-grained modeling of argumentative structures. In contrast, our proposed method, \textbf{ReViewGraph}, explicitly models reviewer-author debates as heterogeneous graphs, captures multi-perspective viewpoints, and performs structured reasoning over interaction relations using graph neural networks.

% \begin{figure*}[t]
%     \centering
%     \includegraphics[width=\linewidth]{review_graph_framework.pdf}
%     \caption{The workflow of ReViewGraph. Reviewer1, Reviewer2, and Reviewer3 represent three regular reviewer agents. Nodes are typed by shape: circles denote paper topic nodes, pentagrams denote evaluation dimension nodes (e.g., Methodological Novelty), squares denote reviewer agents (with different colors distinguishing different agents), and triangles represent the author agent. }
%     \label{fig:ReViewGraph}
% \end{figure*}

\begin{figure*}[t]
    \centering
    \includegraphics[width=\linewidth]{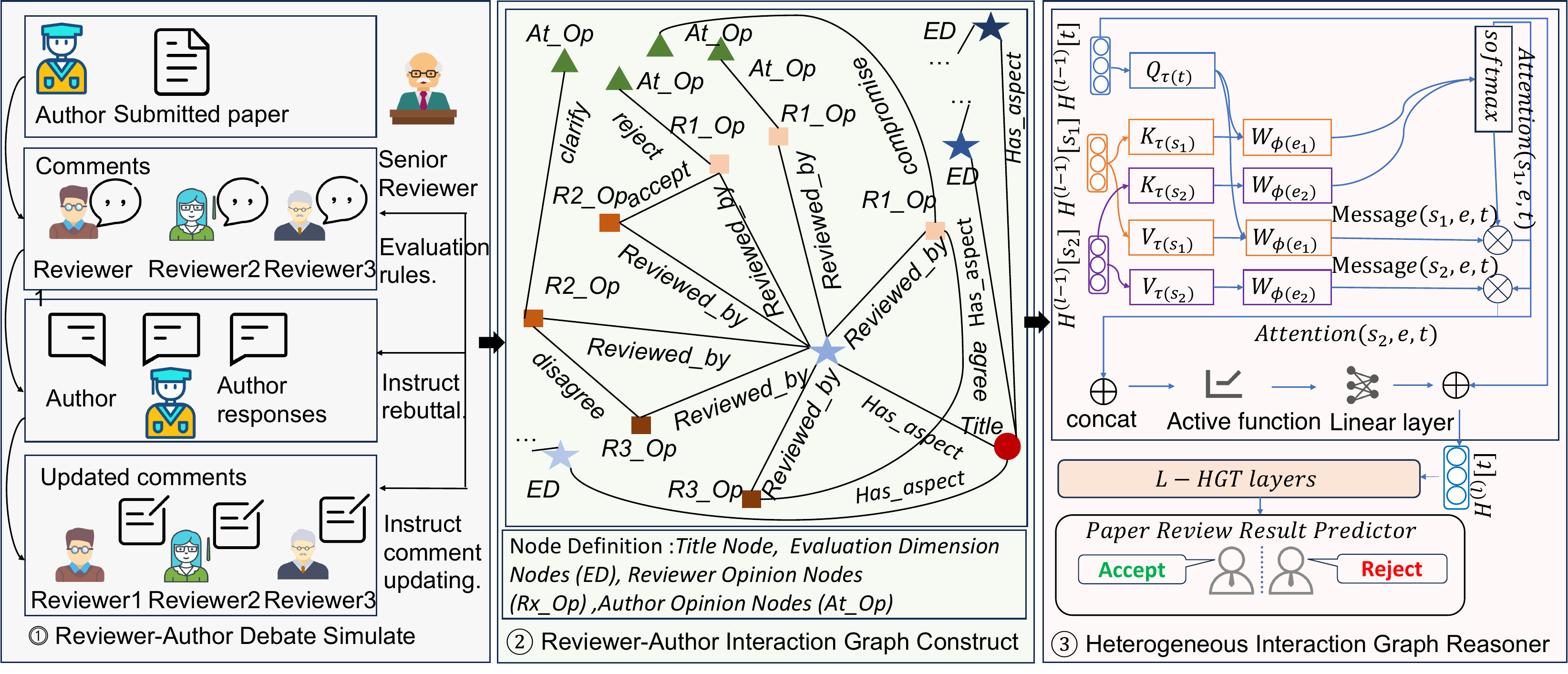}
    \caption{The workflow of ReViewGraph. Reviewer1, Reviewer2, and Reviewer3 represent three regular reviewer agents. Nodes are typed by shape: circles denote paper topic nodes, pentagrams denote evaluation dimension nodes (e.g., Methodological Novelty), squares denote opinions of the reviewers, and triangles represent the author's opinions. }
    \label{fig:ReViewGraph}
\end{figure*}
\section{The ReViewGraph Framework}
% \subsection{Overview}
As illustrated in Figure~\ref{fig:ReViewGraph}, given a paper $D$, ReViewGraph first employs multi-agent collaboration to simulate multi-round debates between reviewers and the author. Based on the generated debate content, it then constructs a heterogeneous interaction graph and performs structured relational reasoning to predict the final review decision $s$ for the paper. We provide a detailed explanation of ReViewGraph in the following.
\subsection{Multi-agent Reviewer-Author Debate Simulation}

To obtain a fine-grained understanding of the target paper, we propose a multi-agent collaboration framework that simulates the dynamics of real-world reviewer–author interactions. The framework consists of four role-specific agents: three regular reviewer agents, one author agent, and a senior reviewer agent who serves as a meta-level coordinator.

The simulation proceeds in three stages: (1) Initial Review Stage.
Each regular reviewer agent is instantiated using a multimodal LLM capable of processing both textual and visual content. Given a set of review criteria and the full content of the target paper $D$, they are encouraged to first recognize the paper’s strengths and substantive contributions, providing positive feedback where appropriate. At the same time, they are expected to raise concerns, highlight potential weaknesses, and offer critical analysis, especially for sections that appear ambiguous or underspecified. This stage ensures a comprehensive and balanced evaluation of the paper’s strengths and limitations.  (2) Author Rebuttal Stage. The senior reviewer agent then prompts the author agent to generate a point-by-point response to the reviewers’ feedback. The author is instructed to clarify any misunderstandings, answer specific technical questions, and defend the paper's contributions where challenged. This stage aims to simulate the rebuttal phase commonly found in peer review, where authors have the opportunity to address regular reviewers' concerns. (3) Re-evaluation Stage.
After receiving the author's rebuttal, the senior reviewer agent reminds the three regular reviewer agents to re-express or refine their opinions in light of the new information. Regular reviewers are encouraged to reassess their initial judgments, revise their critiques, or reaffirm their positions based on the clarified understanding.

This multi-round dialogue process enables the system to capture nuanced argumentative structures and simulate realistic peer review behaviors, laying the foundation for downstream review decisions.

\subsection{Heterogeneous Debate Graph Construction}
To represent the complex reasoning and interaction dynamics of the peer review process, we construct a heterogeneous debate graph from simulated reviewer-author interactions. Following the standard formulation of heterogeneous information networks~\cite{hgt}, we define the graph as $G=\{\mathcal{V},\mathcal{E},\mathcal{A},\mathcal{R}\}$, where $\mathcal{V}$ is the set of nodes, $\mathcal{E}$ is the set of edges, $\mathcal{A}$ is the set of node types, and $\mathcal{R}$ is the set of edge types. Each node $v \in \mathcal{V}$ and $e \in \mathcal{E}$ is associated with a type via node and edge type mapping functions: $\psi(v): V \rightarrow \mathcal{A}$, $\eta(e): E \rightarrow \mathcal{R}$. Each edge corresponds to a \textbf{meta-relation} in the form $<\psi(s),\eta(e),\psi(t)>$, where $s$ and $t$ are source and target nodes.

\subsubsection{Nodes.} $A = \{ \textit{Title},\ \textit{EvaluationDimension} ,\ \textit{ReviewerOpin}$\\$\textit{ion},\ \textit{AuthorOpinion} \}
$, instantiated as follows: (1) Title Node ($\psi(v) = \textit{Title}$): It represents the paper being reviewed, with its content set to full paper title (e.g., \textit{``Automatic Paper Reviewing with Heterogeneous Graph Reasoning over LLM-Simulated Reviewer-Author Debates.''}); (2) Evaluation Dimension Nodes ($\psi(v) = \textit{EvaluationDimension}$): They represent critical dimensions of academic review. To capture a holistic evaluation of a paper's quality, we focus on four core dimensions that are frequently emphasized in real-world peer reviews. These aspects reflect both the intellectual merit and presentational quality of a submission. We define four dimensions in this work: \textit{Methodological Novelty}, \textit{Experimental Completeness}, \textit{Motivation Clarity}, \textit{Writing Fluency}; (3) Reviewer Opinion Nodes ($\psi(v) = \textit{ReviewerOpinion}$): Each reviewer agent produces a set of comments, where each opinion in their comments is represented as a separate node. (4) Author Opinion Nodes ($\psi(v) = \textit{AuthorOpinion}$): The author agent responds to the reviewers with their opinions, which are also modeled as individual nodes.

\subsubsection{Meta-Relations.} We define four types of edges (i.e., relations) $\mathcal{R}$, each corresponding to a meaningful discourse connection. The resulting meta-relations $<\psi(s),\eta(e),\psi(t)>$ include: 
    (1) Paper–Dimension Relations: $<\textit{Title},\textit{has\_aspect},\textit{EvaluationDimension}>$, it represents that each paper is reviewed under several standard dimensions.
    (2) Dimension–Opinion Relations: $<\textit{ReviewerOpinion},\textit{reviewed\_by},\textit{EvaluationDimension}>$\\. It means that each review opinion belongs to a specific evaluation dimension.
    (3) Inter-Reviewer Argumentative Relations: $<\textit{Reviewer} \textit{Opinion},\textit{r},\textit{ReviewerOpinion}>, r \in \{agree, disagree, complement, progressive, independe$\\$nt\}$. These relations reflect viewpoint interactions among reviewers over shared dimensions.
    (4) Reviewer–Author Interaction Relations:
$<\textit{ReviewerOpinion},\textit{r},\textit{AuthorOpinion}$\\$>, r \in \{accept, reject, clarify, compromise, extend, ne$\\$utral\}.$ These edges represent how authors respond to reviewer critiques.
\subsubsection{Graph Instantiation.} To instantiate the graph with the aforementioned meta-relations, we need to extract the relationships among reviewers' opinions as well as those between reviewers' and authors' opinions from their interactions. Additionally, it is necessary to categorize the evaluation dimensions to which each reviewer opinion belongs. To achieve this, we first present the simulated dialogue between reviewers and the author, then prompt it to identify pairs of opinion statements and the relation that holds between them. For instance, if Reviewer A challenges a point raised by Reviewer B, the extracted triplet would reflect this disagreement relation between their respective opinions. Similarly, when an author responds to a reviewer’s critique, the extracted triplet captures the response relation between the author’s and the reviewer’s views.
Next, we further classify each opinion in the triplets into its corresponding Evaluation Dimension (e.g., \textit{Methodological Novelty}, \textit{Experimental Completeness}, \textit{Motivation Clarity}, \textit{Writing Fluency}). This is also achieved via in-context prompting, where the LLM is guided to assign a category to each opinion statement based on its content and argumentative context.

Through this pipeline, we construct a final heterogeneous debate graph that captures multi-perspective opinions and their semantic relationships within reviewer–author debates.
\subsection{Reviewer-Author Debate Graph Reasoning}
Given the reviewer-author debate graph constructed in the previous section, we adopt a Heterogeneous Graph Transformer (HGT) to perform reasoning over the structured interactions between reviewers and authors. Then, we can obtain contextualized representations of all nodes in the constructed heterogeneous graph, which can then be used for downstream tasks such as review result prediction. 

Formally, for a target node $t \in V$, HGT learns its representation by performing heterogeneous mutual attention, heterogeneous message passing, and target-specific aggregation over each source node $s$ connected to $t$ through a relation type (edge). HGT usually consists of $L$ stacked layers, which means the output of the $l$-th HGT layer, denoted as $H^{(l)}$, is fed as input into the next layer. Then the final nodes are represented as $H^{(L)}$.

\subsubsection{Heterogeneous Mutual Attention.} The Heterogeneous Mutual Attention mechanism determines the importance (i.e., weights) of neighboring nodes. For a target node $t$, its neighboring nodes $s \in N(t)$ may be connected via different meta-relations, represented as $<\psi(s),\eta(e),\psi(t)>$. Similar to the vanilla Transformer architecture, HGT maps the target node $t$  to a query vector and each source node $s$ to a key vector, using their dot product to compute attention scores. To model the diverse distributions of meta-relations, HGT further adopts relation-specific projection weights, allowing it to distinguish and effectively capture the semantics of different types of meta-relations. Specifically, in the attention mechanism of HGT, the weight matrices are decomposed into three components corresponding to the projections of the source node $s$, the edge $e$, and the target node $t$. Formally, for each neighboring node $s$ of $t$, the attention mechanism computes multi-head attention scores, concatenates them, and applies the softmax function for normalization:
\begin{align}
    \textbf{ATN}(s, e, t) = \sigma_{\forall s \in \mathcal{N}(t)} [\text{concat}_{i \in [1,Z]} ( \mathrm{attn^{{h}_i}}(s,e,t) ] \\
    \mathrm{attn^{{h}_i}} = (K^i(s)W^{attn}_{\eta(e)}Q^i(t)) \cdot \frac{\mu_{<\psi(s),\eta(e),\psi(t)>}}{\sqrt{d}}
\end{align}

where $Z$ is the number of attention head, and $\sigma$ denotes the Softmax function. For the $i$-th attention head $h_i$, the key of the source node and the query of the target node are first interacted via a relation-specific projection matrix $W_{\eta(e)}$, scaled by a prior weight $\mu_{<\psi(s),\eta(e),\psi(t)>}$ representing importance for different meta relation triplets, and then divided by $\sqrt{d}$, following the standard Transformer scaling. Specifically, $\psi(s)$-type source node $s$ is projected into $K^i(s) \in \mathbb{R}^{d_h}$, and the $\psi(t)$-type target node $t$ is projected into $Q^i(t) \in \mathbb{R}^{d_h}$, where $d_h = d/h$ denotes the dimensionality of each attention head. $W^{attn}_{\eta(e)}$ denotes the edge-based matrix for each edge type $\eta(e)$. 
\subsubsection{Heterogeneous Message Passing.} This mechanism generates the information (i.e., messages) transmitted from neighboring nodes. Specifically, for $(s,e,t)$, the multi-head message can be calculated as:
\begin{align}
    \textbf{MSG}(s,e,t) &= \text{concat}_{i \in [1,Z]}mesg^{h_i}(s,e,t)\\
    mesg^{h_i}(s,e,t) &= \text{Linear}_{\psi(s)}^{i}(H^{(l-1)}[s])W_{\eta(e)}^{mesg}
\end{align}
where $\psi(s)$-type source node $s$ is projected into $i$-th message vector with the linear layer $\text{Linear}_{\psi(s)}^{i}$, and $W_{\eta(e)}^{mesg}$ is used to incorporate the edge dependency. Finally, all messages from $Z$ heads are concatenated as the final message representation for each source node. 
\subsubsection{Target-Specific Aggregation.} Once the attention weights and the messages from the source nodes of the target node $t$ are computed, the representation of $t$ at the $l$-th layer is updated as follows:
\begin{align}
        H^{(l)}[t] &= \text{Linear}_{\psi{(t)}}(\lambda\hat{H}^{(l)}[t]) + H^{(l-1)}[t]\\
        \hat{H}^{(l)}[t] &= \oplus_{\forall s \in \mathcal{N}(t)} \mathrm{\textbf{ATN}}(s, e, t) \cdot \mathrm{\textbf{MSG}}(s, e, t)
\end{align}
where $\oplus$ denotes the weighted summation operation over all source nodes $\forall s \in \mathcal{N}(t)$ of the target node $t$, $\lambda$ serves as a rescaling factor.

\subsubsection{Review Result Prediction with HGT}
After obtaining the vector representation of each node in the heterogeneous debate graph between reviewers and authors, we first apply mean pooling over node embeddings grouped by their types. Since $\psi: V \rightarrow \mathcal{A}$ is a mapping from nodes $v \in V$ to their corresponding node types $a \in \mathcal{A}$, where $\mathcal{A}$ denotes the set of all possible node types. Then, the pooled representation for each node type $a \in \mathcal{A}$ is computed as:
\begin{equation}
    h_a = \frac{1}{|V_a|} \sum_{\substack{v \in V, \psi(v) = a}} H^{(l)}[t]
\end{equation}
where $V_a = \{ v \in V \mid \psi(v) = a \}$ denotes the set of nodes of type $a$. Next, we concatenate the pooled representations of all node types:
\begin{equation}
    h_{\text{concat}} = {||}_{\substack{a \in \mathcal{A}}} h_a
\end{equation}
where $||$ denotes vector concatenation. Finally, the concatenated vector is fed into a two-layer feedforward neural network to predict the final review decision (accept/reject):
\begin{equation}
    \hat{y} = \text{Softmax}\left(W_2 \cdot \text{ReLU}\left(W_1 \cdot h_{\text{concat}} + b_1\right) + b_2\right)
\end{equation}
where $W_1$ and $W_1$ are learnable parameters, and $b_1$ and $b_2$ are biases.

\section{Experiments}

\subsection{Experimental Settings}
\subsubsection{Task.} Following prior works~\cite{agent-review,zhu2025deepreviewimprovingllmbasedpaper,weng2025cycleresearcher,grapheval}, we use the full paper as input and aim to predict the final review decision. The decision $s$ is selected from a predefined set of review outcomes: \textit{accept} or \textit{reject}.

\subsubsection{Datasets and Evaluation Metrics.} To validate the effectiveness of ReViewGraph, we crawled submission data and review outcomes from ICLR 2023, 2024, and 2025 available on OpenReview. The statistics of the collected data are shown in Table~\ref{tab:dataset_statistics_compact}. To thoroughly assess how well the evaluation methods align with human reviewers, we report each method’s performance in terms of Accuracy, Macro Precision, Macro Recall, and Macro F1 score.

\subsubsection{Baselines and Implementation Details.} We compare our proposed framework against several representative baselines that leverage LLMs for automatic paper review: (1) ICL-based Method~\cite{BrownMRSKDNSSAA20}: Directly prompts a pre-trained LLM using in-context examples to predict the review decision based on the full paper content. (2) CoT-based Method~\cite{Wei0SBIXCLZ22}: Encourages the LLM to perform chain-of-thought reasoning before generating the final review outcome. (3) AI-Scientist~\cite{ai-scientist}: Redefines the review process as a multi-turn, long-context dialogue among multiple roles. (4) CycleReviewer~\cite{weng2025cycleresearcher}: Fine-tunes an open-source LLM using domain-specific peer review data. (5) DeepReview~\cite{zhu2025deepreviewimprovingllmbasedpaper}: Extends CycleReviewer by modeling multi-step reviewer reasoning and training on generated rationales. (6) GraphEval~\cite{grapheval}: Segments paper abstracts into discrete opinion sentences using LLM prompting, connects them via similarity-based edges to construct a graph, and applies graph-based reasoning to predict the final review decision. 
The implementation details of ReViewGraph and the baselines are provided in the source code repository.

\begin{table}[t]
\setlength{\tabcolsep}{4pt}
\centering
\small
\begin{tabular}{cl|l|rrr}
\hline
\textbf{Year} & \textbf{Mode} & \textbf{Accept (A-o / A-p / A-s)} & \textbf{Reject} & \textbf{Total} \\
\hline
\multirow{3}{*}{2023} 
& Train & 532 (42 / 349 / 141) & 419 & 951 \\
& Val   & 75 (6 / 49 / 20)     & 59  & 134 \\
& Test  & 155 (13 / 101 / 41)  & 121 & 276 \\
\hline
\multirow{3}{*}{2024} 
& Train & 583 (42 / 361 / 180) & 429 & 1012 \\
& Val   & 82 (6 / 51 / 25)     & 61  & 143 \\
& Test  & 169 (12 / 104 / 53)  & 123 & 292 \\
\hline
\multirow{3}{*}{2025} 
& Train & 712 (107 / 394 / 211) & 631 & 1343 \\
& Val   & 101 (15 / 56 / 30)    & 90  & 191 \\
& Test  & 207 (32 / 114 / 61)   & 181 & 388 \\
\hline
\end{tabular}
\caption{Dataset statistics. Accept numbers are shown with their subcategory breakdown: Oral (A-o), Poster (A-p), Spotlight (A-s).}
\label{tab:dataset_statistics_compact}
\end{table}

\begin{table*}[ht]
\centering
\setlength{\tabcolsep}{5pt}
\begin{tabular}{lllllllllllll}
\hline
\multicolumn{1}{l}{\textbf{Datasets}}                     & \multicolumn{4}{c}{\textbf{ICLR 2023 Papers}}        & \multicolumn{4}{c}{\textbf{ICLR 2024 Papers}} & \multicolumn{4}{c}{\textbf{ICLR 2025 Papers}}        \\
\hline
\multicolumn{1}{l}{\textbf{Method\textbackslash{}Metric}} & \textbf{Acc} & \textbf{P} & \textbf{R} & \textbf{F1} & \textbf{Acc}  & \textbf{P}  & \textbf{R} & \textbf{F1} & \textbf{Acc} & \textbf{P} & \textbf{R} & \textbf{F1} \\
\hline
{ICL-based Method}                                     & 58.33             & 68.75          &         52.66   &     42.44        &     61.30          &  70.69 &      54.39       &   46.65           &       56.70     &       71.74    &   53.66    &43.04   \\
CoT-based Method     &          57.97    &     65.04       &      52.33      &  42.25  &        \underline{61.64}       &       \underline{71.35}      & \underline{54.80 } &     47.38        &   57.47           &      \underline{75.06}      &       54.45     &       44.24      \\
{AI-Scientist}      &       59.06      &     70.64       &     \underline{53.49 }      &      44.03       &  60.48   &     67.32        &  53.69 &     45.65        &       58.51      &     \textbf{78.13 }      &  \underline{55.53}          &      49.95       \\
{GraphEval}  & 50.00    & 47.08  &  47.51  &  46.35   & 48.97 &    44.23       &  45.30 & 43.75 &    46.91 &  44.25  &  45.38   & 43.13    \\
{CycleReviewer-8B}                                 &51.33              &66.67            &25.85            &37.25             &47.83               &66.67             &21.12   &32.08             &49.05              & 55.42           &23.35            &32.86             \\
{CycleReviewer-70B}                                &61.23              &\textbf{82.43}            &39.35            &53.28             &57.73               &71.30             &45.56   &55.60             &63.05              &72.22            &50.24            &\underline{59.26 }            \\
{DeepReview-14B-Std}                         &61.23             &\underline{73.08 }           &49.03            &\underline{58.69 }            &59.93               & 69.70            &54.44   &\underline{61.13}             &\underline{63.77}              &64.29            &47.06            &54.34             \\
\hline
\textbf{ReViewGraph}   &\textbf{70.29}	&	69.85&   \textbf{69.92} & \textbf{69.89}  &  \textbf{66.10} &   65.48 & \textbf{65.73}  & \textbf{65.54}  & \textbf{71.65}	& 72.04   &\textbf{72.01}	& \textbf{71.65}           \\
\hline
\end{tabular}
\caption{Performance Comparison between our ReViewGraph and Other Methods (Acc: Accuracy, P: Macro Precision, R: Macro Recall, F1: Macro F1 Score). The \textbf{bold} number indicates the best performance, while the \underline{underlined} number represents the second-best.}
\label{tab:main_results}
\end{table*}

\begin{table}[ht]
\centering
\small
\begin{tabular}{lcccc}
\hline
\multicolumn{5}{c}{\textbf{ICLR 2023 Papers}} \\
\hline
\textbf{Method\textbackslash{}Metric} & \textbf{Acc} & \textbf{P} & \textbf{R} & \textbf{F1} \\
\hline
\textbf{ReViewGraph} & \textbf{70.29} & \textbf{69.85} & \textbf{69.92} & \textbf{69.89} \\
-- w/o Title & 67.03 & 67.98 & 68.02 & 67.03 \\
-- w/o Eval & 69.20 & 69.63 & 69.86 & 69.17 \\
-- w/o RAR & 69.57 & 69.37 & 69.64 & 69.37 \\
-- w/o IRR & 68.12 & 67.99 & 68.26 & 67.95 \\
-- w/o Hetero & 68.12 & 68.48 & 68.71 & 68.07 \\
\hline
\multicolumn{5}{c}{\textbf{ICLR 2024 Papers}} \\
\hline
\textbf{ReViewGraph} & \textbf{66.10} & 65.48 & 65.73 & \textbf{65.54} \\
-- w/o Title & 64.73 & \textbf{65.78} & \textbf{65.99} & 64.71 \\
-- w/o Eval & 65.75 & 65.60 & \textbf{65.99} & 65.48 \\
-- w/o RAR & 65.75 & 65.25 & 65.55 & 65.28 \\
-- w/o IRR & 65.75 & 65.41 & 65.77 & 65.39 \\
-- w/o Hetero & 65.75 & 65.04 & 65.21 & 65.10 \\
\hline
\multicolumn{5}{c}{\textbf{ICLR 2025 Papers}} \\
\hline
\textbf{ReViewGraph} & \textbf{71.65} & \textbf{72.04} & \textbf{72.01} & \textbf{71.65} \\
-- w/o Title & 66.49 & 69.30 & 67.59 & 66.02 \\
-- w/o Eval & 68.30 & 69.28 & 68.90 & 68.24 \\
-- w/o RAR & 70.36 & 70.71 & 70.70 & 70.36 \\
-- w/o IRR & 69.59 & 70.19 & 70.04 & 69.57 \\
-- w/o Hetero & 70.62 & 71.41 & 71.15 & 70.59 \\
\hline
\end{tabular}
\caption{Results of ablation study.}
\label{tab:ablation_study}
\end{table}
\subsection{Experimental Results}
\subsubsection{Overall Performance.}
As shown in Table~\ref{tab:main_results}, our proposed ReViewGraph consistently outperforms all baseline methods on the ICLR 2023, 2024, and 2025 datasets across four key metrics. Notably, ReViewGraph achieves its strongest performance on the ICLR 2025 dataset, with all metrics exceeding 70 and an average relative improvement of 15.73\% over competitive baselines. To assess statistical significance, we performed two-sample T-tests on the accuracy and F1 scores of ReViewGraph and the best-performing baseline, CycleReviewer-70B. The resulting p-values were 0.0192 and 0.0067, respectively, both below the 0.05 threshold, confirming that the improvements are statistically significant. We further analyze the sources of ReViewGraph’s performance gains in the following sections.

First, compared to prompt-based methods such as ICL-based and CoT-based methods, ReViewGraph demonstrates a substantial performance gain, especially in Accuracy and Macro-F1, with an average improvement exceeding 10 percentage points. This performance gap arises because prompt-based methods rely solely on paper content, limiting their ability to deeply understand and evaluate the review context. In contrast, ReViewGraph employs a multi-agent framework to simulate multi-turn reviewer–author debates, producing rich, structured, and multi-perspective opinion content. 

Secondly, although AI-Scientist also models multi-role dialogues, ReViewGraph goes a step further by explicitly extracting semantic relations between viewpoints (e.g., acceptance, rejection, clarification, compromise). This enhances the semantic precision of interaction modeling, making the structural reasoning more logical and interpretable. Third, ReViewGraph outperforms fine-tuned LLM-based baselines such as CycleReviewer and DeepReview. While CycleReviewer-70B demonstrates strong precision, its recall and overall consistency lag behind. For example, on the ICLR 2025 dataset, ReViewGraph achieves a Macro F1 score of 71.65, outperforming CycleReviewer-70B by more than 12 percentage points. Importantly, unlike these baselines that require resource-intensive fine-tuning, ReViewGraph operates without any LLM parameter updates. By leveraging a heterogeneous graph structure and applying Transformer-based reasoning, our method achieves greater generalizability, training efficiency, and inference controllability. Finally, compared to GraphEval, which constructs graphs based on sentence-level similarity within abstracts, ReViewGraph builds a more semantically expressive and structurally rich heterogeneous graph. It extracts explicit, labeled relations from full-text reviewer–author interactions, leading to higher representational fidelity and improved reasoning accuracy.

\subsubsection{Ablation Analysis.}
We conducted ablation studies on three datasets to evaluate the impact of key components in our heterogeneous reviewer–author opinion graph. Specifically, we removed each of the following elements in turn: the paper title nodes (w/o Title), the evaluation dimension nodes (w/o Eval), the edges representing reviewer–author interactions (w/o RAR), and the edges capturing relations among reviewers’ opinions (w/o IRR). Additionally, we replaced the heterogeneous graph structure with a homogeneous one by collapsing all node and edge types into a single type (w/o Hetero). In this homogeneous graph, all edges are uniformly labeled as \textit{connected}, and nodes are not distinguished by type. This setup allows us to assess the benefits of explicitly modeling heterogeneity in opinion interactions.

As reported in Table~\ref{tab:ablation_study}, removing the paper title nodes (w/o Title) results in the most significant performance drop, indicating the importance of explicitly modeling the target submission. Excluding the evaluation dimension nodes (w/o Eval) also leads to a noticeable decline, confirming that capturing multiple review criteria enhances the model's discriminative capacity. Ablations that remove reviewer–author interaction edges (w/o RAR) or inter-reviewer relational edges (w/o IRR) both lead to moderate decreases, highlighting the value of modeling detailed argumentative and logical relations. Finally, replacing the heterogeneous graph with a homogeneous graph structure (w/o Hetero), where node and edge types are collapsed, further impairs performance, demonstrating the effectiveness of explicitly modeling heterogeneity in reviewer–author debates.

\begin{figure}[t]
    \centering
    \begin{subfigure}[b]{\linewidth}
        \centering
        \includegraphics[width=\linewidth]{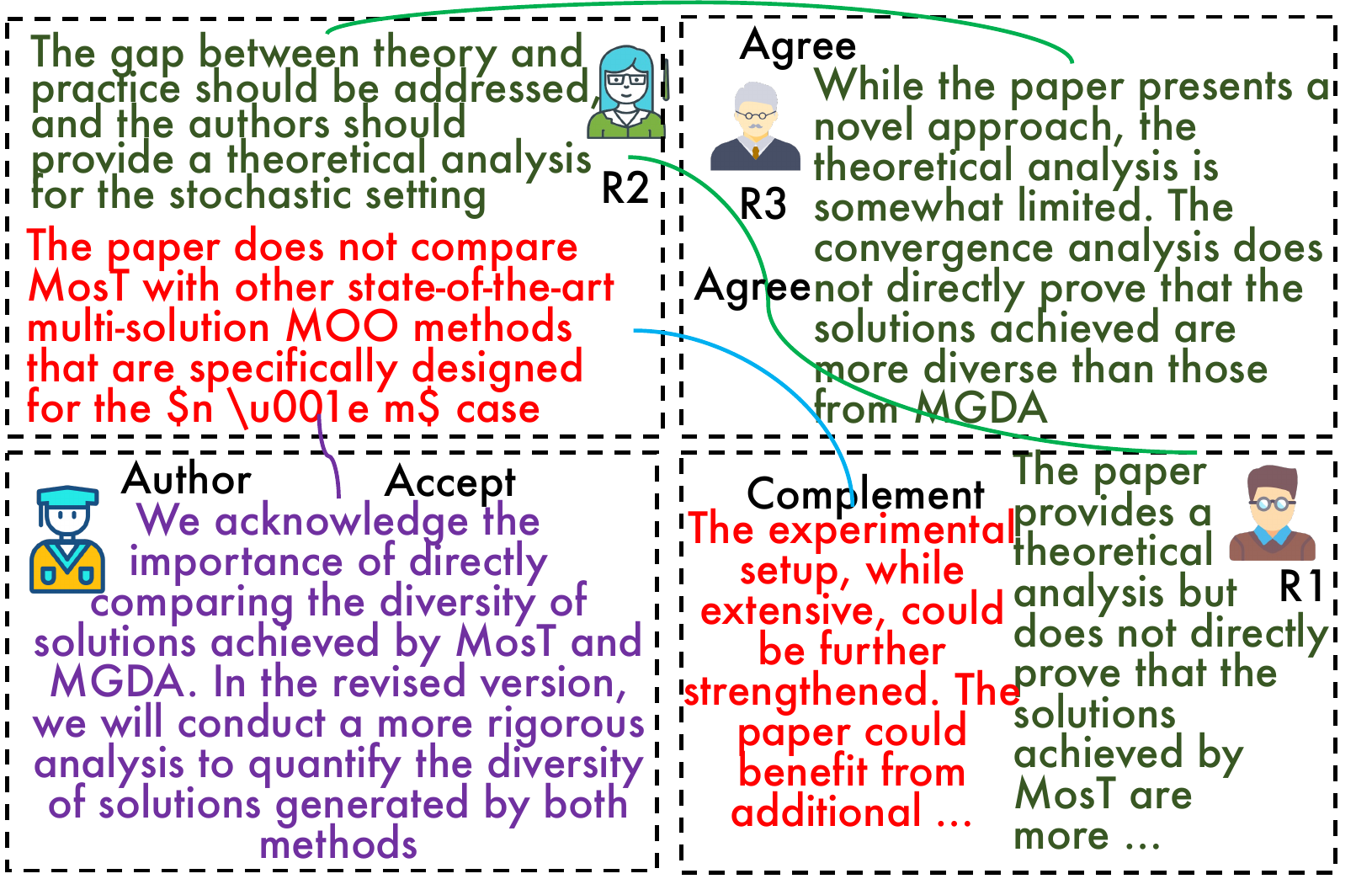}
        \caption{Case 1: Correct rejection prediction by ReViewGraph.}
        \label{fig:case_study_reject}
    \end{subfigure}
    \hfill
    \begin{subfigure}[b]{\linewidth}
        \centering
        \includegraphics[width=\linewidth]{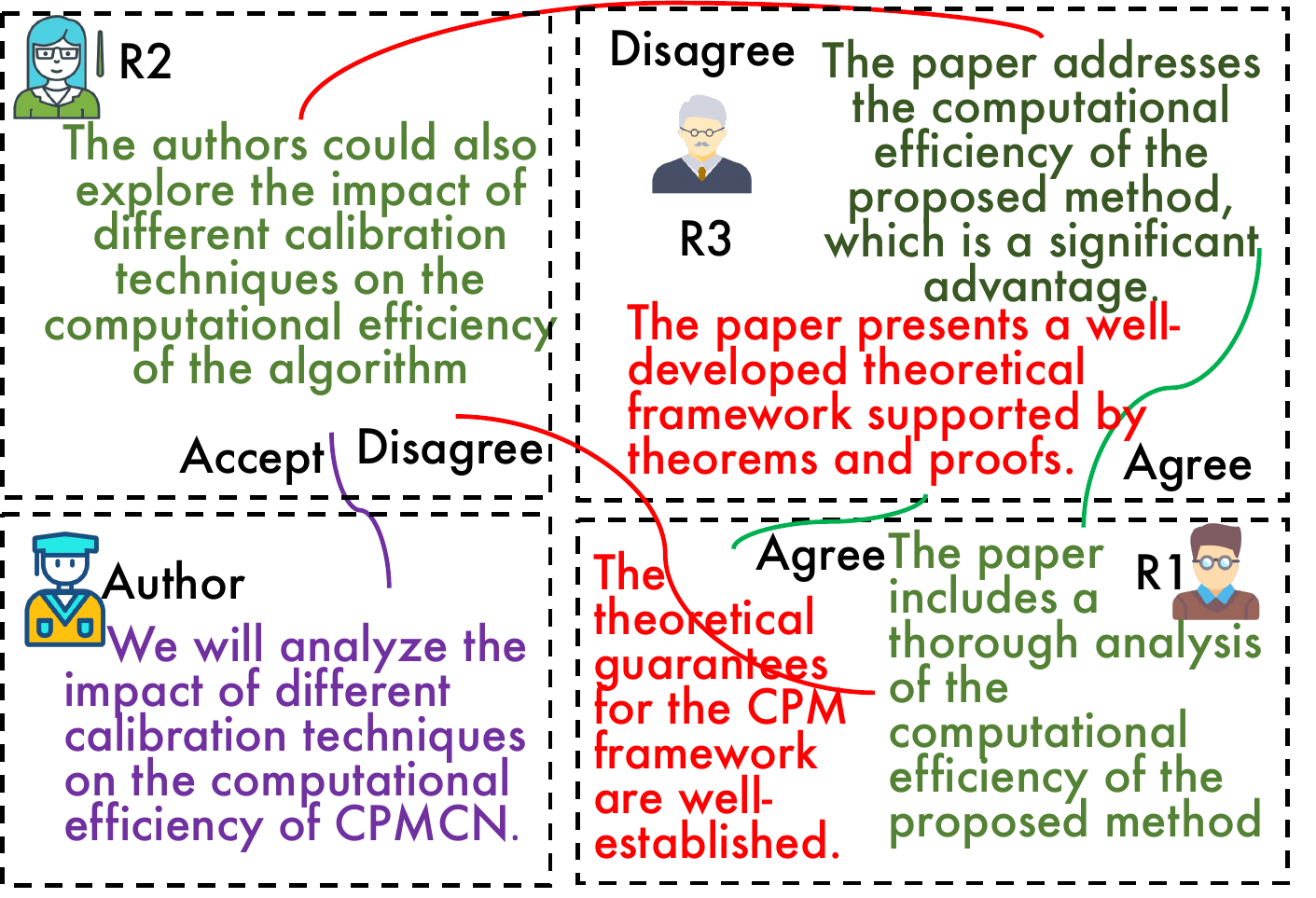}
        \caption{Case 2: Correct acceptance prediction by ReViewGraph.}
        \label{fig:case_study_accept}
    \end{subfigure}
    \caption{Two representative cases of ReViewGraph.}
    \label{fig:case_study_combined}
\end{figure}
\subsubsection{Case Study.} To analyze the advantages of our method, we present two representative cases as follows:
\textbf{Case 1: Correct Rejection Prediction Amid Subtle Negative Consensus.}
In the first case shown in Figure~\ref{fig:case_study_reject}, the ground-truth decision was \textit{Reject}. While the paper proposed a novel approach, multiple reviewers expressed concerns about the lack of rigorous theoretical analysis and inadequate comparisons with state-of-the-art baselines. Reviewer 1 noted the absence of evidence demonstrating the solution's diversity. Reviewer 2 emphasized the theoretical gap in stochastic settings and the lack of comparison with methods designed for the $n \gg m$ scenario. Reviewer 3 echoed similar concerns regarding limited convergence analysis and experimental comparisons. Although the authors acknowledged these issues and promised improvements, no concrete resolutions were provided. ReViewGraph successfully captured the negative consensus across multiple evaluation dimensions (e.g., theoretical rigor, baseline comparison) by modeling reviewer–reviewer agreement and reviewer–author interactions in a structured heterogeneous graph. The system correctly identified the dominant negative stance despite superficially polite language or minor positive comments, and inferred the rejection outcome. In contrast, methods such as ICL, CoT prompting, and LLM-based dialogue agents misclassified the case as \textit{Accept}, likely due to over-reliance on isolated affirmative signals.
\textbf{Case 2: Correct Acceptance Prediction Despite Isolated Criticism.}
In another instance shown in Figure~\ref{fig:case_study_accept}, the ground-truth decision was \textit{Accept}. Reviewer 2 suggested further exploration of the impact of calibration techniques on computational efficiency. However, both Reviewer 1 and Reviewer 3 explicitly stated that the paper adequately addressed computational concerns and praised its theoretical guarantees. Reviewer 2 ultimately agreed with the decision to accept. The authors also acknowledged the suggestion and expressed willingness to expand the analysis in future revisions. While several baseline models incorrectly predicted a \textit{Reject} decision, possibly misinterpreting Reviewer 2’s suggestion as a critical flaw, our model accurately inferred that Reviewer 2’s comment was an isolated and non-decisive concern. By modeling the majority–minority stance across reviewers and the consistency between evaluation dimensions and final decisions, our method avoided overweighing minority dissent. 

% These case studies show that our approach captures structured reviewer interactions and evaluation alignment, enabling more robust and context-aware predictions.

\section{Conclusion}
We proposed \textbf{ReViewGraph}, a novel framework for automatic paper reviewing that leveraged heterogeneous graphs constructed from simulated reviewer–author interactions. By modeling multi-agent debates, we extracted structured opinion relations among reviewers and authors and represented them as a heterogeneous graph with semantically typed nodes and edges. A heterogeneous graph Transformer was then applied to reason over this structure and predict final review outcomes. Through comprehensive experiments on three ICLR datasets, ReViewGraph consistently outperformed strong prompt-based, fine-tuned, and graph-based baselines across multiple evaluation metrics. 
Notably, our method achieved a relative improvement of 15.73\% in Macro F1 score over the best-performing baseline, while requiring no LLM fine-tuning. Ablation studies further validated the importance of each graph component, particularly the modeling of evaluation dimensions and inter-agent relations. Case studies also demonstrated that ReViewGraph was able to correctly interpret nuanced reviewer dynamics, such as subtle consensus or isolated dissent, which were often misclassified by other methods.
% Our method improves Macro-F1 by 15.73\% over the strongest baseline. Ablations verify the necessity of each graph component, and case studies show that ReViewGraph reliably captures nuanced reviewer dynamics often misclassified by prior methods.
These findings highlighted that explicitly modeling reviewer–author interactions and discourse-level semantics led to more robust, interpretable, and context-aware automated reviewing. Overall, ReViewGraph offered a scalable and effective framework for augmenting peer review with structured LLM reasoning, pointing toward future directions in trustworthy AI-assisted scientific evaluation.

\section{Acknowledgments}
We would like to thank the anonymous reviewers
for their valuable comments and suggestions to
improve this paper. Min Yang was supported by National Key Research and Development Program of China (2024YFF0908201), National Natural Science Foundation of China (Grant No. 62376262), the Natural Science Foundation of Guangdong Province of China (2024A1515030166, 2025B1515020032). Xian Wei is supported by the General Program of Shanghai Natural Science Foundation (No.24ZR1419800, No.23ZR1419300), the National Natural Science Foundation of China (No.42130112, No.42371479), the Science and Technology Commission of Shanghai Municipality (No.22DZ2229004), and Shanghai Frontiers Science Center of Molecule Intelligent Syntheses. Shiwen Ni was supported by GuangDong Basic and Applied Basic Research Foundation (2023A1515110718 and 2024A1515012003), China Postdoctoral Science Foundation (2024M753398), Postdoctoral Fellowship Program of CPSF (GZC20232873) and Shenzhen Science and Technology Program  (JCYJ20250604182917023).

\bibliography{aaai2026}

@article{ai-scientist,
  author       = {Chris Lu and
                  Cong Lu and
                  Robert Tjarko Lange and
                  Jakob N. Foerster and
                  Jeff Clune and
                  David Ha},
  title        = {The {AI} Scientist: Towards Fully Automated Open-Ended Scientific
                  Discovery},
  journal      = {CoRR},
  volume       = {abs/2408.06292},
  year         = {2024},
  url          = {https://doi.org/10.48550/arXiv.2408.06292},
  doi          = {10.48550/ARXIV.2408.06292},
  eprinttype    = {arXiv},
  eprint       = {2408.06292},
}

@inproceedings{agent-review,
  author       = {Yiqiao Jin and
                  Qinlin Zhao and
                  Yiyang Wang and
                  Hao Chen and
                  Kaijie Zhu and
                  Yijia Xiao and
                  Jindong Wang},
  title        = {AgentReview: Exploring Peer Review Dynamics with {LLM} Agents},
  booktitle    = {Proceedings of the 2024 Conference on Empirical Methods in Natural
                  Language Processing, {EMNLP} 2024, Miami, FL, USA, November 12-16,
                  2024},
  pages        = {1208--1226},
  publisher    = {Association for Computational Linguistics},
  year         = {2024},
}

@inproceedings{weng2025cycleresearcher,
title={CycleResearcher: Improving Automated Research via Automated Review},
author={Yixuan Weng and Minjun Zhu and Guangsheng Bao and Hongbo Zhang and Jindong Wang and Yue Zhang and Linyi Yang},
booktitle={The Thirteenth International Conference on Learning Representations},
year={2025},
url={https://openreview.net/forum?id=bjcsVLoHYs}
}

@misc{zhu2025deepreviewimprovingllmbasedpaper,
      title={DeepReview: Improving LLM-based Paper Review with Human-like Deep Thinking Process}, 
      author={Minjun Zhu and Yixuan Weng and Linyi Yang and Yue Zhang},
      year={2025},
      eprint={2503.08569},
      archivePrefix={arXiv},
      primaryClass={cs.CL},
      url={https://arxiv.org/abs/2503.08569}, 
}

@misc{alberts2008reviewing,
  title={Reviewing peer review},
  author={Alberts, Bruce and Hanson, Brooks and Kelner, Katrina L},
  journal={Science},
  volume={321},
  number={5885},
  pages={15--15},
  year={2008},
  publisher={American Association for the Advancement of Science}
}

@article{drozdz2024peer,
  title={The peer review process: past, present, and future},
  author={Drozdz, John A and Ladomery, Michael R},
  journal={British Journal of Biomedical Science},
  volume={81},
  pages={12054},
  year={2024},
  publisher={Frontiers Media SA}
}

@article{StelmakhSSD21,
  author       = {Ivan Stelmakh and
                  Nihar B. Shah and
                  Aarti Singh and
                  Hal Daum{\'{e}} III},
  title        = {Prior and Prejudice: The Novice Reviewers' Bias against Resubmissions
                  in Conference Peer Review},
  journal      = {Proc. {ACM} Hum. Comput. Interact.},
  volume       = {5},
  number       = {{CSCW1}},
  pages        = {75:1--75:17},
  year         = {2021},
}

@article{fairness-peer-review,
  author       = {Jiayao Zhang and
                  Hongming Zhang and
                  Zhun Deng and
                  Dan Roth},
  title        = {Investigating Fairness Disparities in Peer Review: {A} Language Model
                  Enhanced Approach},
  journal      = {CoRR},
  volume       = {abs/2211.06398},
  year         = {2022},
  url          = {https://doi.org/10.48550/arXiv.2211.06398},
  doi          = {10.48550/ARXIV.2211.06398},
  eprinttype    = {arXiv},
  eprint       = {2211.06398},
}

@article{llama,
  author       = {Hugo Touvron and
                  Thibaut Lavril and
                  Gautier Izacard and
                  Xavier Martinet and
                  Marie{-}Anne Lachaux and
                  Timoth{\'{e}}e Lacroix and
                  Baptiste Rozi{\`{e}}re and
                  Naman Goyal and
                  Eric Hambro and
                  Faisal Azhar and
                  Aur{\'{e}}lien Rodriguez and
                  Armand Joulin and
                  Edouard Grave and
                  Guillaume Lample},
  title        = {LLaMA: Open and Efficient Foundation Language Models},
  journal      = {CoRR},
  volume       = {abs/2302.13971},
  year         = {2023},
  url          = {https://doi.org/10.48550/arXiv.2302.13971},
  doi          = {10.48550/ARXIV.2302.13971},
  eprinttype    = {arXiv},
  eprint       = {2302.13971},
}

@article{qwen,
  author       = {Jinze Bai and
                  Shuai Bai and
                  Yunfei Chu and
                  Zeyu Cui and
                  Kai Dang and
                  Xiaodong Deng and
                  Yang Fan and
                  Wenbin Ge and
                  Yu Han and
                  Fei Huang and
                  Binyuan Hui and
                  Luo Ji and
                  Mei Li and
                  Junyang Lin and
                  Runji Lin and
                  Dayiheng Liu and
                  Gao Liu and
                  Chengqiang Lu and
                  Keming Lu and
                  Jianxin Ma and
                  Rui Men and
                  Xingzhang Ren and
                  Xuancheng Ren and
                  Chuanqi Tan and
                  Sinan Tan and
                  Jianhong Tu and
                  Peng Wang and
                  Shijie Wang and
                  Wei Wang and
                  Shengguang Wu and
                  Benfeng Xu and
                  Jin Xu and
                  An Yang and
                  Hao Yang and
                  Jian Yang and
                  Shusheng Yang and
                  Yang Yao and
                  Bowen Yu and
                  Hongyi Yuan and
                  Zheng Yuan and
                  Jianwei Zhang and
                  Xingxuan Zhang and
                  Yichang Zhang and
                  Zhenru Zhang and
                  Chang Zhou and
                  Jingren Zhou and
                  Xiaohuan Zhou and
                  Tianhang Zhu},
  title        = {Qwen Technical Report},
  journal      = {CoRR},
  volume       = {abs/2309.16609},
  year         = {2023},
  url          = {https://doi.org/10.48550/arXiv.2309.16609},
  doi          = {10.48550/ARXIV.2309.16609},
  eprinttype    = {arXiv},
  eprint       = {2309.16609},
}

@article{reviewMT,
  author       = {Cheng Tan and
                  Dongxin Lyu and
                  Siyuan Li and
                  Zhangyang Gao and
                  Jingxuan Wei and
                  Siqi Ma and
                  Zicheng Liu and
                  Stan Z. Li},
  title        = {Peer Review as {A} Multi-Turn and Long-Context Dialogue with Role-Based
                  Interactions},
  journal      = {CoRR},
  volume       = {abs/2406.05688},
  year         = {2024},
  url          = {https://doi.org/10.48550/arXiv.2406.05688},
  doi          = {10.48550/ARXIV.2406.05688},
  eprinttype    = {arXiv},
  eprint       = {2406.05688},
}

@inproceedings{ZhouC024,
  author       = {Ruiyang Zhou and
                  Lu Chen and
                  Kai Yu},

  title        = {Is {LLM} a Reliable Reviewer? {A} Comprehensive Evaluation of {LLM}
                  on Automatic Paper Reviewing Tasks},
  booktitle    = {Proceedings of the 2024 Joint International Conference on Computational
                  Linguistics, Language Resources and Evaluation, {LREC/COLING} 2024,
                  20-25 May, 2024, Torino, Italy},
  pages        = {9340--9351},
  publisher    = {{ELRA} and {ICCL}},
  year         = {2024},
}

@inproceedings{grapheval,
  author       = {Tao Feng and
                  Yihang Sun and
                  Jiaxuan You},
  title        = {GraphEval: {A} Lightweight Graph-Based {LLM} Framework for Idea Evaluation},
  booktitle    = {The Thirteenth International Conference on Learning Representations,
                  {ICLR} 2025, Singapore, April 24-28, 2025},
  publisher    = {OpenReview.net},
  year         = {2025},
}

@inproceedings{ErricaSSB25,
  author       = {Federico Errica and
                  Davide Sanvito and
                  Giuseppe Siracusano and
                  Roberto Bifulco},
  title        = {What Did {I} Do Wrong? Quantifying LLMs' Sensitivity and Consistency
                  to Prompt Engineering},
  booktitle    = {Proceedings of the 2025 Conference of the Nations of the Americas
                  Chapter of the Association for Computational Linguistics: Human Language
                  Technologies, {NAACL} 2025 - Volume 1: Long Papers, Albuquerque, New
                  Mexico, USA, April 29 - May 4, 2025},
  pages        = {1543--1558},
  publisher    = {Association for Computational Linguistics},
  year         = {2025},
}

@article{HuangYMZFWCPFQL25,
  author       = {Lei Huang and
                  Weijiang Yu and
                  Weitao Ma and
                  Weihong Zhong and
                  Zhangyin Feng and
                  Haotian Wang and
                  Qianglong Chen and
                  Weihua Peng and
                  Xiaocheng Feng and
                  Bing Qin and
                  Ting Liu},
  title        = {A Survey on Hallucination in Large Language Models: Principles, Taxonomy,
                  Challenges, and Open Questions},
  journal      = {{ACM} Trans. Inf. Syst.},
  volume       = {43},
  number       = {2},
  pages        = {42:1--42:55},
  year         = {2025},
}

@inproceedings{KangADZKHS18,
  author       = {Dongyeop Kang and
                  Waleed Ammar and
                  Bhavana Dalvi and
                  Madeleine van Zuylen and
                  Sebastian Kohlmeier and
                  Eduard H. Hovy and
                  Roy Schwartz},
  title        = {A Dataset of Peer Reviews (PeerRead): Collection, Insights and {NLP}
                  Applications},
  booktitle    = {Proceedings of the 2018 Conference of the North American Chapter of
                  the Association for Computational Linguistics: Human Language Technologies,
                  {NAACL-HLT} 2018, New Orleans, Louisiana, USA, June 1-6, 2018, Volume
                  1 (Long Papers)},
  pages        = {1647--1661},
  publisher    = {Association for Computational Linguistics},
  year         = {2018},
}

@inproceedings{GhosalVEB19,
  author       = {Tirthankar Ghosal and
                  Rajeev Verma and
                  Asif Ekbal and
                  Pushpak Bhattacharyya},
  title        = {DeepSentiPeer: Harnessing Sentiment in Review Texts to Recommend Peer
                  Review Decisions},
  booktitle    = {Proceedings of the 57th Conference of the Association for Computational
                  Linguistics, {ACL} 2019, Florence, Italy, July 28- August 2, 2019,
                  Volume 1: Long Papers},
  pages        = {1120--1130},
  publisher    = {Association for Computational Linguistics},
  year         = {2019},
}

@inproceedings{BartoliLMT16,
  author       = {Alberto Bartoli and
                  Andrea De Lorenzo and
                  Eric Medvet and
                  Fabiano Tarlao},
  title        = {Your Paper has been Accepted, Rejected, or Whatever: Automatic Generation
                  of Scientific Paper Reviews},
  booktitle    = {Availability, Reliability, and Security in Information Systems - {IFIP}
                  {WG} 8.4, 8.9, {TC} 5 International Cross-Domain Conference, {CD-ARES}
                  2016, and Workshop on Privacy Aware Machine Learning for Health Data
                  Science, {PAML} 2016, Salzburg, Austria, August 31 - September 2,
                  2016, Proceedings},
  series       = {Lecture Notes in Computer Science},
  volume       = {9817},
  pages        = {19--28},
  publisher    = {Springer},
  year = {2016},

}

@inproceedings{Nagata19,
  author       = {Ryo Nagata},
  title        = {Toward a Task of Feedback Comment Generation for Writing Learning},
  booktitle    = {Proceedings of the 2019 Conference on Empirical Methods in Natural
                  Language Processing and the 9th International Joint Conference on
                  Natural Language Processing, {EMNLP-IJCNLP} 2019, Hong Kong, China,
                  November 3-7, 2019},
  pages        = {3204--3213},
  publisher    = {Association for Computational Linguistics},
  year         = {2019},
}

@inproceedings{WangZHKJR20,
  author       = {Qingyun Wang and
                  Qi Zeng and
                  Lifu Huang and
                  Kevin Knight and
                  Heng Ji and
                  Nazneen Fatema Rajani},
  title        = {ReviewRobot: Explainable Paper Review Generation based on Knowledge
                  Synthesis},
  booktitle    = {Proceedings of the 13th International Conference on Natural Language
                  Generation, {INLG} 2020, Dublin, Ireland, December 15-18, 2020},
  pages        = {384--397},
  publisher    = {Association for Computational Linguistics},
  year         = {2020},
}

@inproceedings{hgt,
  author       = {Ziniu Hu and
                  Yuxiao Dong and
                  Kuansan Wang and
                  Yizhou Sun},
  title        = {Heterogeneous Graph Transformer},
  booktitle    = {{WWW} '20: The Web Conference 2020, Taipei, Taiwan, April 20-24, 2020},
  pages        = {2704--2710},
  publisher    = {{ACM} / {IW3C2}},
  year         = {2020},

}

@inproceedings{BrownMRSKDNSSAA20,
  author       = {Tom B. Brown and
                  Benjamin Mann and
                  Nick Ryder and
                  Melanie Subbiah and
                  Jared Kaplan and
                  Prafulla Dhariwal and
                  Arvind Neelakantan and
                  Pranav Shyam and
                  Girish Sastry and
                  Amanda Askell and
                  Sandhini Agarwal and
                  Ariel Herbert{-}Voss and
                  Gretchen Krueger and
                  Tom Henighan and
                  Rewon Child and
                  Aditya Ramesh and
                  Daniel M. Ziegler and
                  Jeffrey Wu and
                  Clemens Winter and
                  Christopher Hesse and
                  Mark Chen and
                  Eric Sigler and
                  Mateusz Litwin and
                  Scott Gray and
                  Benjamin Chess and
                  Jack Clark and
                  Christopher Berner and
                  Sam McCandlish and
                  Alec Radford and
                  Ilya Sutskever and
                  Dario Amodei},
  title        = {Language Models are Few-Shot Learners},
  booktitle    = {Advances in Neural Information Processing Systems 33: Annual Conference
                  on Neural Information Processing Systems 2020, NeurIPS 2020, December
                  6-12, 2020, virtual},
  year         = {2020},
}

@inproceedings{Wei0SBIXCLZ22,
  author       = {Jason Wei and
                  Xuezhi Wang and
                  Dale Schuurmans and
                  Maarten Bosma and
                  Brian Ichter and
                  Fei Xia and
                  Ed H. Chi and
                  Quoc V. Le and
                  Denny Zhou},
  title        = {Chain-of-Thought Prompting Elicits Reasoning in Large Language Models},
  booktitle    = {Advances in Neural Information Processing Systems 35: Annual Conference
                  on Neural Information Processing Systems 2022, NeurIPS 2022, New Orleans,
                  LA, USA, November 28 - December 9, 2022},
  year         = {2022},
}

@inproceedings{KingmaB14,
  author       = {Diederik P. Kingma and
                  Jimmy Ba},
  title        = {Adam: {A} Method for Stochastic Optimization},
  booktitle    = {3rd International Conference on Learning Representations, {ICLR} 2015,
                  San Diego, CA, USA, May 7-9, 2015, Conference Track Proceedings},
  year         = {2015},
}
% \clearpage
% \input{input/reproduction_check_list}
\clearpage
\section{Appendix}
% \subsection{Example of Constructed Heterogeneous Graph}
\subsection{Implementation Details.} 
\subsubsection{ReViewGraph.}
Given the early stage of research in this area and the absence of a standardized benchmark, we construct a new dataset to support the simulation of multimodal reviewer-author debates. Inspired by prior work, we collect papers from OpenReview\footnote{\url{https://openreview.net}}spanning ICLR 2023, 2024, and 2025. For each paper, we retain the main text and references, while excluding appendices to focus on the core scientific content. We then apply the MinerU\footnote{\url{https://github.com/HIT-SCIR/MinerU}}  tool to extract structured multimodal information, including textual paragraphs, figures, and tables. This curated dataset provides the necessary input for downstream prediction and will be publicly released upon publication to facilitate future research in this area.

To enable multimodal processing in the multi-agent simulation, we adopt Qwen2.5-VL-72B-Instruct\footnote{\url{https://huggingface.co/Qwen/Qwen2.5-VL-72B-Instruct}} as the backbone model for all agents, due to its ability to process both textual and visual inputs. To construct a heterogeneous debate graph, we first simulate reviewer-author interactions using a multi-agent collaboration framework. The prompt used to generate these simulated interactions is presented in Table~\ref{tab:multi_agent_prompt}. Based on the generated interaction text, we then employ GPT-4.1-mini to extract opinion triples, which include various types of meta-relations defined in the main body of the paper. The prompt used for triple extraction is shown in Table~\ref{tab:prompt_triple_extraction}. Table~\ref{tab:triple_extraction_example_rejected} and Table~\ref{tab:triple_extraction_example_accepted} provide example outputs for the two types of triple extraction results. To further associate the extracted opinions with specific evaluation dimensions, we continue to prompt GPT-4.1-mini using the template illustrated in Table~\ref{tab:argumentative_sentence_classification}, enabling opinion classification and alignment with the corresponding evaluation dimensions. 

For structured reasoning over reviewer-author interactions, we design a heterogeneous graph and apply a two-layer Heterogeneous Graph Transformer (HGT) with a total of three layers. Each layer uses multi-head attention, and the hidden dimension is set to 128. To initialize node and edge representations, we use Qwen3-Embedding\footnote{\url{https://huggingface.co/Qwen/Qwen3-Embedding-8B}}. The model is trained for up to 100 epochs with a batch size of 32, using early stopping based on validation F1 score. Optimization is performed using the Adam optimizer~\cite{KingmaB14} with a learning rate of $1 \times 10^{-4}$.
Our code and data are provided in the supplementary materials.

\subsubsection{Baselines.} To comprehensively evaluate the effectiveness of our proposed approach, we compare it against a diverse set of baseline methods spanning different prompting paradigms and reasoning strategies.

Table~\ref{tab:prompt-icl-based-method} and Table~\ref{tab:prompt_cot_rewriting} respectively summarize the prompt construction and implementation details of the ICL-based~\cite{BrownMRSKDNSSAA20} and CoT-based methods~\cite{Wei0SBIXCLZ22}. Table~\ref{tab:ai-scientist_paper_review} and Table~\ref{tab:ai-scientist_paper_review_reflection_prompt} present the prompt templates used by the AI-Scientist~\cite{ai-scientist} method for paper reviewing and review aggregation, respectively. All baselines are run using consistent inference settings (e.g., temperature, max tokens) unless otherwise stated.

For the GraphEval~\cite{grapheval} baseline, we follow the original implementation to preprocess our data and apply the same model configuration for training and evaluation\footnote{\url{https://github.com/ulab-uiuc/GraphEval}}. Specifically, we adopt a Graph Convolutional Network (GCN) as the backbone, configured as a two-layer weighted GNN with a hidden dimension of 64. The model is trained for up to 100 epochs with a batch size of 32, using the Adam optimizer.

For DeepReview~\cite{zhu2025deepreviewimprovingllmbasedpaper} and CycleReview~\cite{weng2025cycleresearcher}, we directly invoke the publicly released models, DeepReviewer-14B\footnote{\url{https://huggingface.co/WestlakeNLP/DeepReviewer-14B}}, CycleReviewer-ML-LLaMA-3.1-8B\footnote{\url{https://huggingface.co/WestlakeNLP/CycleReviewer-ML-Llama-3.1-8B}}, and CycleReviewer-LLaMA-3.1-70B\footnote{\url{https://huggingface.co/WestlakeNLP/CycleReviewer-Llama-3.1-70B}}, without any fine-tuning or additional adaptation. This ensures a fair and reproducible baseline comparison, highlighting the benefits of our explicitly structured multi-agent debate framework.

\subsection{Evaluation Metric Definitions and Calculation Details.}
We report standard classification metrics including Accuracy (Acc), Macro Precision (P), Macro Recall (R), and Macro F1 Score (F1). These metrics are computed based on the predictions $\hat{y}_i$ and the gold labels $y_i$ for each instance $i$ in the evaluation set. Formally:

\begin{equation}
    \text{Accuracy} = \frac{1}{N} \sum_{i=1}^{N} \mathbb{I}(y_i = \hat{y}_i)
\end{equation}

\begin{equation}
    \text{Macro Precision} = \frac{1}{C} \sum_{c=1}^{C} \frac{TP_c}{TP_c + FP_c}
\end{equation}

\begin{equation}
    \text{Macro Recall} = \frac{1}{C} \sum_{c=1}^{C} \frac{TP_c}{TP_c + FN_c}
\end{equation}

\begin{equation}
    \text{Macro F1} = \frac{1}{C} \sum_{c=1}^{C} \frac{2 \cdot \text{Precision}_c \cdot \text{Recall}_c}{\text{Precision}_c + \text{Recall}_c}
\end{equation}

\noindent
Here, $C$ is the number of classes (in our case, 2: \texttt{accept} and \texttt{reject}), and $TP_c$, $FP_c$, and $FN_c$ denote the number of true positives, false positives, and false negatives for class $c$, respectively.

\noindent
The indicator function $\mathbb{I}(y_i = \hat{y}_i)$ returns 1 if the predicted label matches the gold label, and 0 otherwise. For our binary classification setting, this corresponds to correctly or incorrectly predicting whether a paper should be \texttt{accepted} or \texttt{rejected}.

% \subsubsection{LLM-simulated Reviewer-Author Debate}

\begin{table*}[ht]
\footnotesize
\begin{tcolorbox}[colframe=black!80, colback=gray!10, fonttitle=\bfseries\large, coltitle=black, boxrule=0.5mm, arc=5mm, auto outer arc, width=\textwidth,toptitle=6pt, bottomtitle=6pt]

\textbf{System:}

You are a participant in the paper review and you need to fully understand the content of the paper.\\

\textbf{User:}

The following is the text content, figures and tables of a research paper.\\

\textbf{Assistant:}

Okay, please provide the content of the research paper.\\

\textbf{User:}

\code{\{\{paper content\}\}}\\

\textbf{Assistant:}

Received the text content of the research paper.\\

\textbf{User:}

\code{\{\{image url\}\}}\\
\code{\{\{image description\}\}}\\

\textbf{Assistant:}

Received the figure of the research paper.\\

\textbf{User:}

\code{\{\{table url\}\}}\\
\code{\{\{table description\}\}}\\

\textbf{Assistant:}

Received the table of the research paper.\\

\textbf{User:}
You are a helpful assistant. Your role:

You are a reviewer. You write peer review of academic papers by evaluating their technical quality, originality, and clarity.

\#\# Review Guidelines

\code{\{\{review guidelines\}\}}\\

\textbf{Assistant:}

Overall rating: 8

Significance and novelty: The paper introduces a novel method, Image2Sphere (I2S)...\\

\textbf{User:}
You are a helpful assistant. Your role:

You are an author. You write research papers and submit them to conferences. During the rebuttal phase, you carefully read the reviews from the reviewers and respond to each of them.

\#\# Author Guidelines

\code{\{\{author guidelines\}\}}\\

\textbf{Assistant:}

Response: We appreciate the detailed feedback from Reviewer 1 and will address each of the points raised...\\
\\
...\\
\\
\textbf{User:}

You are a helpful assistant. Your role:

You are a very knowledgeable and experienced area chair in a top-tier machine learning conference.

You evaluate the reviews provided by reviewers and write metareviews. 

\#\# Area Chair Guidelines

\code{\{\{area chair guidelines\}\}}\\

\textbf{Assistant:}

Score: 8

Summary: The paper introduces Image2Sphere (I2S), a novel method that leverages SO(3)-equivariant layers...

\end{tcolorbox}
\caption{Prompt template for LLM-simulated reviewer-author debate.}
\label{tab:multi_agent_prompt}
\end{table*}

% \subsubsection{Prompt for Triple Extraction}

\begin{table*}[ht]

\footnotesize
\begin{tcolorbox}[colframe=black!80, colback=gray!10, fonttitle=\bfseries\large, coltitle=black, boxrule=0.5mm, arc=5mm, auto outer arc, width=\textwidth,toptitle=6pt, bottomtitle=6pt]
% \hline
\textbf{System:}\\
You are an expert in argument mining and peer review analysis.  

You are given a review file that contains:

- The initial review comments from three reviewers.

- The author’s responses to each reviewer.

- The reviewers' subsequent responses to the author's replies.\\

\textbf{User:}

\#\#\# Initial Review Comments:

\#\#Reviewer 1: \code{\{\{Comment from Reviewer 1\}\}} ...

\#\#Reviewer 2: ...

\#\#\# Author's Responses:

\#\#Author: \code{\{\{Author's response\}\}}

\#\#Author: ...

\#\#\# Reviewers' Responses:

\#\#Reviewer 1:

\code{\{\{Reviewers' response\}\}}

\#\#Reviewer 2: ...\\

\textbf{Assistant:}

Received the review contents.

\textbf{User:}

\#\#\# Task:

Please carefully read the provided text and extract the argumentative relationships in the following two categories:

1. **Reviewer-Author Relationships:**  

Identify the relationships between each reviewer's argument sentence and the corresponding author's response.  

Use the following relation types:  

- Accept: The author fully accepts or agrees with the reviewer's comment.

- Reject: The author disagrees with or does not adopt the reviewer’s comment.  

- Clarify: The author provides additional explanations or clarifications.  

- Compromise: The author partially accepts the reviewer's comment and proposes a middle-ground solution.

- Extend:  The author expands or supplements their response based on the reviewer's comment. 

- Neutral: The author's response does not clearly express an attitude or is neutral.

2. **Inter-Reviewer Relationships:**  

Identify the relationships between argument sentences from different reviewers.  

Use the following relation types:  

- Agree: The reviewers hold consistent viewpoints or mutually support each other's comments.

- Disagree: The reviewers present conflicting viewpoints or directly refute each other.

- Complement: Reviewers' comments are complementary and cover different but related aspects.  

- Progressive: One reviewer's comment builds upon or deepens another’s.  

- Independent: Reviewers’ comments are unrelated and focus on different issues independently.

\#\#\# Output Format:

Please provide the extracted triples in the following format:

- (Reviewer X: [argument sentence], Author: [argument sentence], [relation type])

- (Reviewer X: [argument sentence], Reviewer Y: [argument sentence], [relation type])

\#\#\# Notes:

- ``Argument sentence" refers to a concise sentence that clearly expresses a viewpoint, criticism, suggestion, or justification.

- Please focus on argument sentences and ignore background descriptions or non-argumentative text.

- If no clear relation exists, skip that pair.

\#\#\# Example Output:

\{``Reviewer\_Author\_Relations": [``(Reviewer 1: `The experiment settings lack sufficient diversity to fully validate the generalizability of the proposed method.', Author: `We have added new experiments on additional datasets from different domains to enhance diversity and support generalization.', Accept)",

...],
    
  ``Inter\_Reviewer\_Relations": [``(Reviewer 1: `The proposed model demonstrates significant novelty in its hierarchical reasoning structure.', Reviewer 2: `The hierarchical reasoning structure introduced is indeed novel and well-motivated.', Agree)",
    
  ...]\}

Please extract all relevant triples from the provided text. Each triple must directly quote the original sentences without any paraphrasing, summarizing, or abbreviation. Ensure that the wording is fully preserved as in the original text. Follow the example and provide the final output in a single JSON object. Please strive to cover the full diversity of relationship types as comprehensively as possible. The extracted relationships should be complete, diverse, and richly capture the different types of interactions present in the text.\\
\end{tcolorbox}
\caption{Prompt template for triple extraction.}
\label{tab:prompt_triple_extraction}
\end{table*}

\begin{table*}[ht]
\footnotesize

\begin{tcolorbox}[colframe=black!80, colback=gray!10, fonttitle=\bfseries\large, coltitle=black, boxrule=0.5mm, arc=5mm, auto outer arc, width=\textwidth,toptitle=6pt, bottomtitle=6pt]

  ``\textbf{Reviewer\_Author\_Relations}": [
  
    ``(\textbf{Reviewer 1}: `While the paper presents a novel method, it lacks a thorough comparison with existing techniques, such as PCA, to fully establish its superiority or unique advantages.', \textbf{Author}: `We acknowledge the need for a more thorough comparison between PTA and PCA. In the revised version, we will include a detailed analysis of the differences between PTA and PCA, focusing on their respective strengths and limitations.', Accept)",
    
    ``(\textbf{Reviewer 1}: `The experimental section could be more robust, as it only includes two case studies.', \textbf{Author}: `We agree that additional case studies would further validate the effectiveness and versatility of PTA. In the revised manuscript, we will include more games or scenarios to demonstrate the method's applicability across a broader range of game structures.', Accept)",
    
    ``(\textbf{Reviewer 2}: `While the paper presents a novel method, it lacks a thorough comparison with existing techniques, such as PCA and other game decomposition methods.', \textbf{Author}: `We acknowledge the importance of comparing PTA with existing methods such as PCA and other game decomposition techniques. In the revised version, we will include a detailed comparison section where we analyze the strengths and weaknesses of PTA relative to these methods.', Accept)",
    
    ``(\textbf{Reviewer 2}: `The paper's empirical evaluation is limited to two specific games, which may not be representative of the broader range of games where PTA could be applied.', \textbf{Author}: `To address the concern about the limited scope of our empirical evaluation, we will expand our analysis to include a broader range of games. This will include non-zero-sum games and games with more than two players to demonstrate the versatility and generalizability of PTA.', Accept)",
    
    ``(\textbf{Reviewer 3}: `The paper lacks a thorough discussion of the limitations and potential drawbacks of PTA.', \textbf{Author}: `We will expand our discussion to include the performance of PTA in games with more complex structures and higher dimensions. We will analyze how the method scales with increasing complexity and provide insights into its effectiveness in such scenarios.', Accept)",
    
    ``(\textbf{Reviewer 3}: `The paper lacks a thorough discussion of the limitations and potential drawbacks of PTA.', \textbf{Author}: `We will incorporate a section that examines the impact of noise and uncertainty in the data on PTA's performance. This will include simulations with varying levels of noise to demonstrate the method's robustness and suggest potential strategies for mitigating adverse effects.', Accept)",
    
    ``(\textbf{Reviewer 3}: `The paper does not provide a comparison of PTA with other existing methods for analyzing game structures.', \textbf{Author}: `We will include a comparative analysis of PTA with other decomposition methods like PCA and nonnegative matrix factorization. This will highlight the unique advantages of PTA, such as its ability to capture cyclic structures and strategic trade-offs.', Accept)",],\\

``\textbf{Inter\_Reviewer\_Relations}": [

``(\textbf{Reviewer 1}: `While the paper presents a novel method, it lacks a thorough comparison with existing techniques, such as PCA, to fully establish its superiority or unique advantages.', \textbf{Reviewer 2}: `While the paper presents a novel method, it lacks a thorough comparison with existing techniques, such as PCA and other game decomposition methods.', Agree)",

``(\textbf{Reviewer 1}: `The paper could benefit from a more detailed discussion of the computational complexity and scalability of PTA.', \textbf{Reviewer 2}: `The paper does not provide a detailed discussion of the computational complexity and scalability of PTA, which are important considerations for practical applications.', Agree)",

``(\textbf{Reviewer 2}: `The paper's writing could be improved in terms of clarity and organization.', \textbf{Reviewer 3}: `The paper is well-written and organized, with a clear flow of ideas and a logical structure.', Disagree)",

``(\textbf{Reviewer 3}: `The paper does not provide a comparison of PTA with other existing methods for analyzing game structures.', \textbf{Reviewer 1}: `While the paper presents a novel method, it lacks a thorough comparison with existing techniques, such as PCA, to fully establish its superiority or unique advantages.', Complement)",

``(\textbf{Reviewer 3}: `The paper lacks a thorough discussion of the limitations and potential drawbacks of PTA.', \textbf{Reviewer 1}: `The paper could benefit from a more detailed discussion of the computational complexity and scalability of PTA.', Complement)",

``(\textbf{Reviewer 2}: `The paper is limited to two case studies and should test on more diverse games including non-zero-sum and more players.', \textbf{Reviewer 3}: `The authors will expand empirical evaluation to a broader range of games.', Agree)",

``(\textbf{Reviewer 1}: `The authors should provide a more detailed analysis of the computational complexity and scalability of PTA.', \textbf{Reviewer 3}: `The authors should provide a detailed discussion of the computational complexity and scalability of PTA.', Agree)",

``(\textbf{Reviewer 3}: `The paper is well-written and organized, with a clear flow of ideas and a logical structure.', \textbf{Reviewer 1}: `The paper could benefit from a more detailed discussion of the computational complexity and scalability of PTA.', Independent)",]

\label{tab:example_of_triple_extraction_results_for_rejected_paper}
\end{tcolorbox}
\caption{Example of triple extraction results for one rejected paper.}
\label{tab:triple_extraction_example_rejected}
\end{table*}

\begin{table*}[ht]

\footnotesize
\begin{tcolorbox}[colframe=black!80, colback=gray!10, fonttitle=\bfseries\large, coltitle=black, boxrule=0.5mm, arc=5mm, auto outer arc, width=\textwidth,toptitle=6pt, bottomtitle=6pt]

``\textbf{Reviewer\_Author\_Relations}": [
    
    ``(\textbf{Reviewer 1}: `The study focuses solely on linear regression, which may not generalize to more complex function classes.', \textbf{Author}: `We acknowledge that our current analysis is limited to linear regression. In the revised version, we will include a discussion on the limitations of our findings when applied to non-linear functions. We will also provide insights into how our results might extend to more complex scenarios, such as non-linear models and other function classes.', Accept)",
    
    ``(\textbf{Reviewer 2}: `The paper does not discuss the scalability of their findings to larger models and datasets.', \textbf{Author}: `We plan to conduct experiments with deeper transformers and larger datasets to assess the robustness and practical applicability of our theoretical constructions.', Accept)",
    
    ``(\textbf{Reviewer 3}: `The paper should include a more comprehensive review of related work.', \textbf{Author}: `We will include a more comprehensive review of related work, discussing how our findings build upon and extend previous research in the field.', Accept)",
    
    ``(\textbf{Reviewer 1}: `The authors have not provided specific details on how they plan to extend their theoretical analysis to include a broader range of learning algorithms or how they will explore the boundaries of what learning algorithms can be implemented by transformers.', \textbf{Author}: `The authors have not provided specific details on this.', Neutral)",
    
    ``(\textbf{Reviewer 2}: `While they have mentioned the need to extend their theoretical analysis to include a broader range of learning algorithms, they have not provided a clear plan for how they will achieve this.', \textbf{Author}: `The authors have not provided a clear plan.', Neutral)",
    
    ``(\textbf{Reviewer 2}: `They have not outlined specific steps for exploring the boundaries of what learning algorithms can be implemented by transformers or for conducting a comparative analysis of the computational costs of implementing learning algorithms within transformers versus traditional methods.', \textbf{Author}: `The authors have not outlined specific steps.', Neutral)",
    
    ``(\textbf{Reviewer 3}: `While they have mentioned the need to extend their theoretical analysis to include a broader range of learning algorithms, they have not provided a clear plan for how they will achieve this.', \textbf{Author}: `The authors have not provided a clear plan.', Neutral)",],\\
  
  ``\textbf{Inter\_Reviewer\_Relations}": [
  
    ``(\textbf{Reviewer 1}: `The study focuses solely on linear regression, which may not generalize to more complex function classes.', \textbf{Reviewer 2}: `The study focuses solely on linear regression, which may not generalize to more complex function classes.', Agree)",
    
    ``(\textbf{Reviewer 1}: `The paper could benefit from a more comprehensive analysis of the algorithmic properties of in-context learners.', \textbf{Reviewer 2}: `The paper lacks a detailed analysis of the computational efficiency and resource requirements of the implemented learning algorithms within transformers.', Complement)",
    
    ``(\textbf{Reviewer 2}: `The paper does not adequately discuss the relationship between their findings and existing work on in-context learning and meta-learning.', \textbf{Reviewer 3}: `The paper should include a more comprehensive review of related work and discuss how the authors' findings build upon and extend previous research in the field.', Agree)",
    
    ``(\textbf{Reviewer 1}: `The paper does not fully explore the practical implications of the findings for real-world applications.', \textbf{Reviewer 2}: `The authors have agreed to provide a more detailed discussion of the practical implications of their findings, including potential applications and ethical considerations.', Progressive)",
    
    ``(\textbf{Reviewer 1}: `The paper does not fully explore the practical implications of the findings for real-world applications.', \textbf{Reviewer 3}: `The paper does not provide a detailed discussion of the implications of its findings for the broader field of machine learning and artificial intelligence.', Agree)",
    
    ``(\textbf{Reviewer 2}: `The paper could benefit from a discussion on the potential challenges and limitations of applying the insights gained from this study to real-world problems.', \textbf{Reviewer 3}: `The authors should discuss how their results contribute to the understanding of in-context learning and its potential applications in real-world scenarios.', Agree)",
    
    ``(\textbf{Reviewer 2}: `The paper does not evaluate the performance of in-context learners on real-world datasets.', \textbf{Reviewer 3}: `The authors should conduct additional experiments with different datasets and training regimes to validate the robustness of the empirical results.', Complement)"]

\end{tcolorbox}
\caption{Example of triple extraction results for one accepted paper.}
\label{tab:triple_extraction_example_accepted}
\end{table*}

% \subsubsection{Prompt for Argumentative Sentence Classification}
\begin{table*}[ht]

\begin{tcolorbox}[colframe=black!80, colback=gray!10, fonttitle=\bfseries\large, coltitle=black, boxrule=0.5mm, arc=5mm, auto outer arc, width=\textwidth,toptitle=6pt, bottomtitle=6pt]

\textbf{System:}\\
\#\#\# Task:

Your task is to classify the given comment into one of the following categories based on their primary evaluation focus:

1. Methodological Novelty — Is the comment evaluating whether the proposed method is novel, creative, or technically original?

2. Motivation Clarity — Is the comment assessing whether the motivation or problem statement is clear and compelling?

3. Experimental Completeness — Is the comment about the quality, completeness, or reliability of the experiments and empirical evaluation?

4. Writing Fluency — Is the comment about the writing quality, fluency, or readability of the paper?

Return your answer as JSON in the format: {``category": ``category\_name"}\\

\textbf{User:}\\
Comment to classify: \code{\{\{comment\}\}}\\

\end{tcolorbox}
\caption{Prompt template for argumentative sentence classification.}
\label{tab:argumentative_sentence_classification}
\end{table*}
% \subsection{Implementation Details for Baselines}
% \subsubsection{ICL-based Method}
% prompt of it 

% \subsubsection{Prompted LLIM Method}
\begin{table*}[ht]
\begin{tcolorbox}[colframe=black!80, colback=gray!10, fonttitle=\bfseries\large, coltitle=black, boxrule=0.5mm, arc=5mm, auto outer arc, width=\textwidth,toptitle=6pt, bottomtitle=6pt]
\textbf{System:}

You are an AI researcher who is reviewing a paper's content that was submitted to a prestigious ML venue. Be critical and cautious in your decision.

If a paper's content is bad or you are unsure, give it bad scores and reject it!

$<$Instruction$>$

Please evaluate the paper draft based on the following six dimensions: 

\code{\{\{paper evaluation criteria prompt template\}\}}

You will classify the paper into one of the following four categories based on the evaluation:

\code{\{\{paper evaluation decision prompt template\}\}}

Note: The approximate distribution of decisions for papers at this ML venue is as follows: Reject - 61\%; Accept - 39\%. Therefore, you should only accept your paper if you are absolutely sure that it is acceptable; otherwise, you should reject it. Please take this decision distribution into account and make your judgment carefully.

$<$Input$>$

Here is the paper draft to evaluate: Paper content – \{\{content\}\};

$<$ Output$>$

You only need to give an overall score (0-10) and select a review decision. No detailed analysis is required.

The output format should follow these rules:

Overall Score (0-10)= \{score\}

\{one decision from ``Reject", ``Accept"\}

An example of the output:

Overall Score (0-10)= 7

Reject\\

\textbf{User:}

Paper content: \code{\{\{content\}\}}\\

\textbf{Assistant:}

Overall Score (0-10)= 8

Accept\\ 

\end{tcolorbox}
\caption{Prompt template for ICL-based method.}
\label{tab:prompt-icl-based-method}
\end{table*}

% \subsubsection{CoT-based Method}
\begin{table*}[ht]
\begin{tcolorbox}[colframe=black!80, colback=gray!10, fonttitle=\bfseries\large, coltitle=black, boxrule=0.5mm, arc=5mm, auto outer arc, width=\textwidth,toptitle=6pt, bottomtitle=6pt]
\textbf{System:}

$<$Instruction$>$

Please evaluate the paper draft step by step based on the following dimensions. For each step, carefully think through and evaluate the corresponding dimension, and then provide ratings for each dimension (1-10). You must give an overall score (0-10) along with the 6 dimension scores. No detailed analysis is needed, but ensure that your evaluation for each step is based on logical reasoning.

$<$Input$>$

Here is the paper draft to evaluate: Paper content – \code{\{\{content\}\}};

$<$Step 1: Evaluate Novelty$>$

First, evaluate the novelty of the paper.\code{\{\{text for novelty in the paper evaluation criteria prompt template.\}\}}

Novelty Rating (1-10):

$<$Step 2: Evaluate Validity$>$

Next, evaluate the validity of the paper. \code{\{\{text for validity in the paper evaluation criteria prompt template.\}\}}

Validity Rating (1-10):

$<$Step 3: Evaluate Significance$>$

Then, evaluate the significance of the paper. \code{\{\{text for significance in the paper evaluation criteria prompt template.\}\}}

Significance Rating (1-10):

$<$Step 4: Evaluate Rigorousness$>$

Now, evaluate the rigorousness of the paper. \code{\{\{text for rigorousness in the paper evaluation criteria prompt template.\}\}}

Rigorousness Rating (1-10):

$<$Step 5: Evaluate Clarity$>$

Next, evaluate the clarity of the paper. \code{\{\{text for clarity in the paper evaluation criteria prompt template.\}\}}

Clarity Rating (1-10):

$<$Step 6: Evaluate Ethical Considerations$>$

Lastly, evaluate the ethical considerations of the paper. \code{\{\{text for ethnic in the paper evaluation criteria prompt template.\}\}}

Ethical Considerations Rating (1-10):

$<$Step 7: Final Overall Score$>$

After completing all the dimension evaluations, summarize your assessment and give an overall score that reflects the paper’s general quality and performance across all dimensions.

Overall Score (0-10):

$<$Step 8: Final Decision$>$ 

Based on the overall score and individual ratings, choose the most appropriate review decision. Carefully consider how the paper performs in each dimension, and select from the following categories:

\code{\{\{paper evaluation decision prompt template\}\}}

Decision:

Note: The approximate distribution of decisions for papers at this ML venue is as follows: \code{\{\{label distribution of the dataset\}\}}. Therefore, you should only accept your paper if you are absolutely sure that it will be accepted. Please take this decision distribution into account and make your judgment carefully.

$<$Output$>$

The output format should follow these rules:

Overall Score (0-10)= \{score\}

Decision: \{one decision from ``Reject", ``Accept"\}

An example of the output:

Overall Score (0-10): 7

Decision: Reject\\

\textbf{User:}

Paper content:\code{\{\{content\}\}}\\

\textbf{Assistant:}

Overall Score (0-10): 8

Decision: Accept\\
\end{tcolorbox}
\caption{Prompt template for CoT-based method.}
\label{tab:prompt_cot_rewriting}
\end{table*}

% \subsubsection{AI-Scientist}
\begin{table*}[ht]
\begin{tcolorbox}[colframe=black!80, colback=gray!10, fonttitle=\bfseries\large, coltitle=black, boxrule=0.5mm, arc=5mm, auto outer arc, width=\textwidth,toptitle=6pt, bottomtitle=6pt]
\textbf{System:}

$<$Instruction$>$

You are an AI researcher who is reviewing a paper that was submitted to a prestigious ML venue. Be critical and cautious in your decision. If a paper is bad or you are unsure, give it bad scores and reject it.\\

\textbf{User:}

Reviewing a submission: step-by-step

Summarized in one sentence, a review aims to determine whether a submission will bring sufficient value to the community and contribute new knowledge. The process can be broken down into the following main reviewer tasks:

\code{\{\{ICLR\_reviewer\_guidelines\}\}}

Here is the paper you are asked to review: 

``` 

\code{\{\{content\}\}}

```\\

\textbf{Assistant:}

Review 1/N: \{review\_1\}

...\\

\textbf{User:}

Round \code{\{\{current\_round\}\}} / \code{\{\{num\_reflections\}\}}.

In your thoughts, first carefully consider the accuracy and soundness of the review you just created.

Include any other factors that you think are important in evaluating the paper.
Ensure the review is clear and concise, and the JSON is in the correct format.
Do not make things overly complicated.

In the next attempt, try and refine and improve your review.

Stick to the spirit of the original review unless there are glaring issues.

Respond in the same format as before:

THOUGHT:

$<$THOUGHT$>$

REVIEW JSON:

```json

$<$JSON$>$

```

If there is nothing to improve, simply repeat the previous JSON EXACTLY after the thought and include ``I am done" at the end of the thoughts but before the JSON.

ONLY INCLUDE ``I am done" IF YOU ARE MAKING NO MORE CHANGES.

\end{tcolorbox}
\caption{Prompt template for AI-Scientist: Paper Review Step.}
\label{tab:ai-scientist_paper_review}
\end{table*}

\begin{table*}[ht]

\begin{tcolorbox}[colframe=black!80, colback=gray!10, fonttitle=\bfseries\large, coltitle=black, boxrule=0.5mm, arc=5mm, auto outer arc, width=\textwidth,toptitle=6pt, bottomtitle=6pt]
\textbf{System:}

You are an Area Chair at a machine learning conference. You are in charge of meta-reviewing a paper that was reviewed by \{reviewer\_count\} reviewers. Your job is to aggregate the reviews into a single meta-review in the same format. Be critical and cautious in your decision, find consensus, and respect the opinion of all the reviewers.\\

\textbf{User:}

Review 1/N: 

\code{\{\{review\_1\}\}}  

...

Review N/N: 

\code{\{\{review\_N\}\}}  

\code{\{\{ICLR\_reviewer\_guidelines\}\}}

\end{tcolorbox}
\caption{Prompt template for AI-Scientist: Paper Review Ensembling Step.}
\label{tab:ai-scientist_paper_review_reflection_prompt}
\end{table*}

\end{document}